\documentclass[11pt,a4paper]{article}

\usepackage[T2A,T1]{fontenc}
\usepackage[utf8]{inputenc}
\usepackage[ukrainian,english]{babel}
\DeclareUnicodeCharacter{2212}{\ensuremath{-}}
\DeclareUnicodeCharacter{00B7}{\textperiodcentered}
\usepackage[margin=2.5cm]{geometry}
\usepackage{amsmath,amssymb}
\usepackage{booktabs,graphicx,rotating,tabularx}
\usepackage{tikz}
\usetikzlibrary{positioning,arrows.meta,fit}
\usepackage[numbers,sort&compress]{natbib}
\usepackage[unicode,pdfencoding=auto,hidelinks]{hyperref}
\usepackage[capitalise,nameinlink]{cleveref}
\usepackage{orcidlink}

\title{An AI Assistant for Automating Anti-Corruption Expertise of Draft Legislation Based on the TRACE Framework\thanks{Preprint, November 2026.}}
\author{Serhii Zabolotnii\,\orcidlink{0000-0003-0242-2234}$^{1,2,3}$,
Alina Tarasova\,\orcidlink{0009-0007-0285-7816}$^{4}$\\[4pt]
\small $^{1}$\,Department of Information, Multimedia Technologies and Design,\\
\small Cherkasy State Business College, Cherkasy 18028, Ukraine\\
\small $^{2}$\,State Scientific Research Institute of Armament and Military Equipment\\
\small Testing and Certification, Cherkasy, Ukraine\\
\small $^{3}$\,Department of Cybernetics and Applied Mathematics,\\
\small Uzhhorod University, Uzhhorod 88000, Ukraine\\
\small $^{4}$\,Legal Counsel, PJSC ``Finport Technologies Inc.'', Ukraine\\[2pt]
\small \texttt{zabolotnii.serhii@csbc.edu.ua}}
\date{}

\begin{document}
\maketitle

\begin{abstract}
\sloppy
We present an AI assistant for automating anti-corruption expertise of draft legislation based on the TRACE architecture, designed to screen bills for corruption risks: pinpoint provisions an expert would criticise and name the regulatory defect in each. The pipeline combines a deterministic parsing core, model-guided single-provision qualification, rule-based factor assignment, and external human oversight. To evaluate the assistant, we constructed an open benchmark from the published conclusions of Ukraine's National Agency on Corruption Prevention (NAZK), comprising 132 draft laws with 474 model-extracted findings. A system flag agrees with a finding when it targets a compatible provision and an LLM judge rates the risk as the same (strict) or partly the same (lenient). On 100 draft laws, the pipeline matches 0.28 of the findings within its top ten flags using the local Gemma~4~31B model, 0.30 with Sonnet~5.5, and 0.43--0.46 across all flags (non-blind; the frozen blind baseline achieves 0.21, 95\,\% CI 0.16--0.26). Across all runs, most misses occur on provisions that the system located but qualified as a different risk, and none of the 32 legal-collision findings was matched. On 21 acts held out from tuning, the pipeline outperforms a monolithic prompt containing the full methodology under the strict measure, though not under the lenient one, and Opus~5.5 adds nothing over Sonnet. The reported scores reflect agreement with model-extracted findings rated by an LLM judge whose two access routes agree on 73 of 74 labels, without human expert validation.

\end{abstract}


\section{Introduction}\label{sec:intro}

Anti-corruption expertise of draft acts is a statutory procedure in Ukraine carried out by the
National Agency on Corruption Prevention (NAZK). Under the methodology approved by NAZK Order
No.~109/23~\citep{nazk109}, an expert identifies a provision that creates conditions for corruption,
assigns the corresponding corruption-risk factor from a 30-factor taxonomy, and recommends how to
remove it. NAZK conducts this review selectively for drafts submitted to the Verkhovna Rada and the
Cabinet of Ministers and publishes the resulting conclusions. Hundreds of public conclusions are already available, yet the incoming stream of drafts far exceeds what manual review can cover. A practical question follows: can an automated system indicate which provisions of a draft an expert should inspect first and what risk to look for in them?

Prior work relevant to automating anti-corruption expertise has either proposed using AI to flag vague
expressions and regulatory gaps in drafted legislation \citep{araszkiewicz2025} or discussed the promise and risks of AI-based anti-corruption tools
at a conceptual level \citep{kobis2022}. Meanwhile, existing benchmarks for legal language models focus on bar exams, court
decisions, contracts, or question answering
\citep{guha2023,chlapanis2025,ovcharov2026,hendrycks2021}. The closest empirical task we found
estimates the probability that a bill is unconstitutional against constitutional-court rulings
\citep{corazza2025}. To our knowledge, no benchmark evaluates provision by provision whether an automated
system reproduces the published conclusions of a statutory review body.

We constructed such a benchmark from published NAZK conclusions and used it to test a layered
pipeline driven by a local language model. Following the TRACE architecture \citep{zabolotnii2026trace}, the system combines a deterministic core,
retrieval and language-model components, an orchestration layer, and external human oversight. On 100 draft laws, the pipeline with Gemma~4~31B
reproduces 28\,\% of NAZK's findings within its first ten flags, compared with 30\,\% for the cloud model
Sonnet~5.5, and reaches up to 46\,\% across all flags (\cref{tab:test100}); \cref{sec:test100} reports these non-blind estimates alongside the blind score of the frozen initial parser (21\,\%). Most misses happen when the pipeline locates the right provision but qualifies the risk differently than NAZK (\cref{fig:loss}). Compared with a monolithic prompt containing the entire methodology and draft text, the structured pipeline on the same model identifies NAZK's exact risk more often without increasing the number of matched provisions (\cref{sec:stage-a}). Because both the reference findings and the evaluation labels come from language models without human verification, all reported scores measure relative agreement on this benchmark (\cref{sec:limits}).

The paper makes four contributions:
\begin{itemize}
\item \textbf{Benchmark and evaluation protocol.} We release a dataset of 132 draft-to-conclusion pairs (474 findings) together with an address-gated evaluation protocol in which a system flag must first match the provision address before an LLM judge compares the stated risks (\cref{sec:benchmark}, \cref{tab:dataset}).
\item \textbf{Empirical comparison of pipeline structure and model scale.} We evaluate the pipeline on
100 draft laws with local and cloud models and compare it against a monolithic prompt on 21 blind acts (\cref{sec:results}, \cref{tab:test100,tab:stage-a}).
\item \textbf{Stage-by-stage loss diagnosis.} We break down missed findings by pipeline stage and methodology factor to isolate where current models fall short (\cref{sec:loss}, \cref{fig:loss}, \cref{tab:factors}).
\item \textbf{The corruption-risk dial.} We describe a visual interface that places the history of amended provisions and earlier NAZK conclusions next to the system's flags (\cref{sec:dial}).
\end{itemize}

Each stage of the pipeline maps directly onto a layer of the TRACE architecture \citep{zabolotnii2026trace}. The deterministic core L1 parses the draft and normalises provision addresses without calling a language model. In layer L2, classical BM25 retrieval (L2a) selects candidate provisions and relevant NAZK examples, while the language model (L2b) is invoked strictly to qualify one provision at a time through a structured schema, leaving the methodology factor to be derived by deterministic rules. Layer L3 ranks the resulting flags and optionally re-prioritises the top ten in a single call per act. Human supervision L4 remains outside the software boundary, where the NAZK specialist writes the official conclusion. This design directly tests TRACE's principle of computational parsimony: whether a mid-size local model suffices once deterministic code handles document structure and taxonomy mapping. A similar decomposition was previously applied to an assistant for Supreme Court legal positions \citep{zabolotnii2026court}, where two independent expert panels evaluated seven models. The present study provides a second legal instantiation of TRACE and the first evaluated against published conclusions of a state oversight agency. Here the architecture, the reasoning schema, and the amendment parser serve as experimental instruments; our focus is on the benchmark, the empirical measurements, and the analysis of failure modes.

\section{Anti-corruption expertise and related work}\label{sec:background}

The Methodology of anti-corruption expertise approved by NAZK Order No.~109/23 of 18 May 2023
\citep{nazk109} replaced the agency's earlier 2020 guide. It defines corruption-risk factors as
properties of a provision that create conditions for a corruption offence and pairs each factor with
standard recommendations for removing the defect. In a published NAZK conclusion, every finding is anchored to a specific
provision of the draft. Across the conclusions we collected, the text typically summarises the provision first, then introduces the regulatory defect and factor after a contrastive marker (such as ``however'' or ``at the same time''), and closes with a
recommendation; \citet{diorditsa2024} describe the broader administrative procedure in detail. For evaluation design, a key property of a NAZK
conclusion is that it does not claim to exhaust every flaw in a bill, recording only the risks the agency judged material.

Earlier work relevant to automating anti-corruption review approached the problem either from legislative drafting or from an institutional perspective. \citet{araszkiewicz2025} examined how AI can support the principles of proper legislation, including flagging
vague expressions and regulatory gaps such as undefined competences of a public body or missing procedural elements;
such defects resemble several corruption-risk factors of the NAZK methodology, notably broad discretion and the absence of procedural rules. Looking at the institutional framework, \citet{kobis2022} analysed the promise and perils of AI-based
anti-corruption tools and argued that their effect depends on who deploys them and how, with top-down use by governments carrying distinct risks. Neither study, however, tested how closely an automated system reproduces the actual conclusions of an expert agency. More broadly, research on machine learning in administrative law emphasises that advisory tools must leave the legal decision with the human official while making the tool's own performance auditable \citep{coglianese2017,busuioc2021}.

General benchmarks for legal language models cover a wide range of tasks, yet none targets statutory anti-corruption review. LegalBench \citep{guha2023} compiles diverse legal reasoning tasks, CUAD \citep{hendrycks2021} and
LexGLUE \citep{chalkidis2022} test contract extraction and classification, GreekBarBench
\citep{chlapanis2025} grades free-text bar-exam answers with an LLM judge, and UA-Legal-Bench
\citep{ovcharov2026} evaluates tasks drawn from Ukrainian court decisions. Among rule-checking datasets, the
closest analogue is the compliance review of national technical standards \citep{wang2026}. While grading open-ended legal explanations with
an LLM judge has become standard practice \citep{zheng2023}, a judge model's alignment with human specialists requires separate verification; our protocol measures only inter-route reproducibility for the same judge model (\cref{sec:protocol}).

Finally, evaluating an amending bill requires reconstructing how its instructions alter the statute in force. Automatic consolidation of statutes and extraction of amending commands have been studied across several jurisdictions
\citep{etcheverry2024}, including German \citep{prior2025} and Polish
\citep{smywinski2021} statutory law, and our deterministic parser for Ukrainian amending bills builds on the same principles. Once a provision is isolated, we qualify it through schema-guided reasoning, prompts the model to populate fields sequentially from verbatim evidence to the specific risk category (an approach related to structured rule templates studied by \citet{yang2025}). The closest end-to-end legislative task is \citet{corazza2025}, who use a hybrid AI pipeline to estimate whether a bill risks being held unconstitutional, though at the level of the bill rather than provision by provision against agency conclusions.

\section{Benchmark}\label{sec:benchmark}

The benchmark measures what fraction of the findings in a published NAZK conclusion an automated
system reproduces at the same provision and with the same underlying risk. All dataset splits and evaluation
rules were registered prior to the model runs reported in \cref{sec:results}.

\subsection{Corpus}

We assembled the digest of published anti-corruption expertise conclusions from the NAZK
portal, comprising 627 conclusions in total: 230 on draft laws, 391 on draft resolutions of the Cabinet of
Ministers, and 6 on other acts. Of these, 609 conclusions identify at least one corruption-risk factor, while the public feed contains no
conclusions reporting zero factors. For every conclusion on a draft law, we retrieved the corresponding bill text from the Verkhovna Rada portal in the edition active on the conclusion date.
This yielded 132 complete pairs with extracted findings (\cref{sec:gold}), consisting of 131 draft laws and one resolution. The bills in the test split are born-digital
PDFs, whereas the 9 acts in the holdout split are scanned documents recognised with Tesseract, and 23 acts in test100
are scans transcribed for experimental purposes by Sonnet~5.5 rather than a production OCR pipeline.

\subsection{Benchmark findings}\label{sec:gold}

Because a NAZK conclusion is written as narrative prose, comparing it against system flags requires structuring each finding into a discrete record containing the provision address in the draft (e.g.\ ``subpara.~4,
para.~15, Section~I''), the methodology factor, a verbatim quote stating the risk, and the
agency's recommendation. We used the model \texttt{gpt-6.1-sol} under a fixed instruction (prompt version \texttt{gold-v1}) to extract these records across all 132 pairs. Given the conclusion text and the
30-factor taxonomy, the model outputs one structured record per finding along with a confidence tag. Both the extraction script and the resulting records are included in the public release. Because these records were not manually curated, every entry in the dataset carries the status
``draft''. On the 21 acts of the test and holdout splits, an automated check that searched for the extracted quotes inside the bill texts surfaced six discrepancies:
three findings cited a pre-registration draft whose provisions had been dropped before formal registration and were therefore excluded, while three required address corrections where item numbering had shifted between versions. Removing the three obsolete entries leaves 474 evaluated findings out of 477. Throughout the paper, the term ``NAZK findings'' refers to these model-extracted records.

\begin{table}[t]
\centering
\caption{Composition of the benchmark across splits (132 draft--conclusion pairs, 474 model-extracted findings). The \texttt{test100}
split incorporates all 12 \texttt{test} acts alongside 88 additional bills (\texttt{test100: new acts}); \texttt{clean15} is a 15-act subset of \texttt{test100} reserved for human verification. No other overlaps exist among \texttt{dev}, \texttt{test}, \texttt{holdout}, and \texttt{test100}.}
\label{tab:dataset}
\begin{tabular}{@{}lrr@{}}
\toprule
Split & Acts & Findings \\
\midrule
All pairs & 132 & 474 \\
\midrule
dev & 8 & 24 \\
test & 12 & 27 \\
holdout & 9 & 29 \\
test100 & 100 & 364 \\
clean15 & 15 & 57 \\
test100: new acts & 88 & 337 \\
\bottomrule
\end{tabular}

\end{table}

\subsection{Splits}\label{sec:splits}

The dataset partitions mirror the sequence of experiments (\cref{tab:dataset}). Prompts and heuristics were tuned on \texttt{dev} (8 acts). The
Stage~A pipeline was then frozen before evaluating \texttt{test} (12 acts) and \texttt{holdout} (9 acts), making those 21 acts a strictly blind test. Next, the \texttt{test100} split (combining the 12 \texttt{test} acts with
88 newly collected bills, plus a 15-act subset \texttt{clean15} sampled for manual audit) and its evaluation protocol were registered ahead of the model runs. Only the initial pipeline commit frozen prior to registration is blind on \texttt{test100}; subsequent runs incorporate L1 parser repairs informed by error diagnostics on \texttt{test100} and are therefore marked non-blind.

\subsection{Evaluation protocol}\label{sec:protocol}

A system run produces an ordered list of flags, each specifying a provision address, a factor, a risk explanation, and a supporting quote. Evaluation proceeds in two stages: deterministic address matching followed by semantic risk comparison.

\paragraph{Address match and LLM judge.}
A system flag is eligible to match a NAZK finding only if their normalized addresses are compatible (meaning they point to the exact same unit or one contains the other, after resolving superscript indices
Art.~28\textsuperscript{1} $=$ Art.~28-1 and article spans ``Articles 39--41''). Findings that have no address-compatible flag among the system's top $K$ outputs are scored as missed immediately, without invoking the judge.
Whenever one or more compatible flags exist within the top $K$, the judge model (\texttt{gpt-6.1-sol}; prompt in
\cref{app:judge}) receives the full text of the NAZK finding, its risk quote, and all address-compatible flags blinded to system identity.
For each candidate pair, the judge assigns one of three labels: \emph{same risk}, \emph{partly the same} (the flag targets the same provision and a related regulatory flaw, but states the defect too broadly or incompletely), or \emph{different}. A finding is counted as reproduced if at least one of the first $K$ compatible flags achieves the required label. Because NAZK occasionally raises multiple distinct findings against a single long article, matching is many-to-many; the judge evaluates each finding independently, although a single overly broad flag rated \emph{partial} against two findings on the same article can slightly inflate lenient recall.

\paragraph{Metrics, uncertainty, and judge consistency.}
We report two primary metrics at cut-off $K=10$: \emph{strict recall} (``same'') and \emph{lenient recall} (``same + partial''). For runs that emit long candidate lists, we also report recall over all emitted flags ($K=\text{all}$) alongside the mean number of flags per act. Standard precision cannot be computed from public data because NAZK conclusions document only selected material risks rather than exhaustive annotations, so an unmatched system flag is not necessarily a false positive.

Confidence intervals (95\,\%) are obtained via cluster bootstrap resampling at the act level (5,000 replicates) to account for within-act correlation among findings. Pairwise system comparisons use a paired bootstrap over the same acts, accompanied by the win/loss count of acts where one system reproduces more findings than its comparator; differences whose 95\,\% confidence interval lies entirely above zero are typeset in bold. Finally, to verify that the judge itself is stable across API providers, we scored four Stage~A runs independently through both Codex and OpenRouter. Across the 74 findings that triggered a judge call in both routes, the assigned three-way labels agreed in 73 cases (\cref{tab:judge}), confirming high inter-route reproducibility.

\begin{table}[t]
\centering
\caption{Inter-route consistency of the LLM judge (\texttt{gpt-6.1-sol}): independent queries through two API routes assign identical labels on 73 of 74 judged findings. ``Judged'' counts findings that had at least one address-compatible flag in both runs; $\kappa$ denotes three-class Cohen's kappa.}
\label{tab:judge}
\begin{tabular}{@{}lrrrr@{}}
\toprule
Run & Findings & Judged & Label agreement & $\kappa$ \\
\midrule
Gemma, test & 27 & 16 & 15 & 0.91 \\
Gemma, holdout & 29 & 18 & 18 & 1.00 \\
Sonnet, test & 27 & 19 & 19 & 1.00 \\
Sonnet, holdout & 29 & 21 & 21 & 1.00 \\
\midrule
Total & 112 & 74 & 73 & \\
\bottomrule
\end{tabular}

\end{table}

\begin{table}[t]
\centering
\small
\caption{Pre-registered evaluation questions and decision criteria. L1: deterministic parser; L3: list-wise ranking layer; CI: 95\,\% paired-bootstrap confidence interval over acts.}
\label{tab:criteria}
\begin{tabularx}{\textwidth}{@{}p{3.6cm}XX@{}}
\toprule
Question & Design & Positive answer if \\
\midrule
Does pipeline structure gain over a monolithic prompt on the same model? &
TRACE v1 vs monolithic prompt, Gemma~4~31B, blind test and holdout acts pooled (21) &
the lower bound of the 95\,\% CI of the paired difference in ``same + partial'' $>0$ \\
Does L3 prioritisation narrow the gap between the local and the cloud model? &
test100, L1 v1.1 for both models (non-blind) &
the gap with prioritisation is smaller than without it and not distinguishable from zero \\
What does structure give a strong model? &
Sonnet~5.5, test (12), all flags &
recall not lower than the monolithic prompt with at most a third of its flags \\
Where is the ceiling? &
breakdown of missed findings by stage, holdout and test100 &
the category ``location found, risk different'' is the largest in every run \\
\bottomrule
\end{tabularx}
\end{table}

\section{System under test}\label{sec:system}

\begin{figure}[t]
\centering
\begin{tikzpicture}[node distance=4mm and 6mm, every node/.style={font=\small},
  box/.style={draw, rounded corners=2pt, align=center, minimum height=11mm, text width=27mm},
  data/.style={draw, dashed, rounded corners=2pt, align=center, text width=27mm, minimum height=9mm},
  arr/.style={-{Stealth[length=2mm]}}]
\node[data] (in) {Draft act,\\comparison table};
\node[box, right=of in] (l1) {\textbf{L1} parsing\\into provisions,\\addresses};
\node[box, right=of l1] (l2a) {\textbf{L2a} candidates\\+ NAZK examples\\(BM25)};
\node[box, below=8mm of l2a] (l2b) {\textbf{L2b} qualifying\\one provision\\(reasoning schema)};
\node[box, left=of l2b] (l3) {\textbf{L3} factor from\\risk type,\\ranking, top 10};
\node[data, left=of l3] (out) {Flags:\\address, factor,\\risk, quote};
\node[data, below=4mm of l2b, text width=27mm] (bank) {Example bank:\\361 NAZK findings};
\draw[arr] (in) -- (l1); \draw[arr] (l1) -- (l2a); \draw[arr] (l2a) -- (l2b);
\draw[arr] (l2b) -- (l3); \draw[arr] (l3) -- (out); \draw[arr] (bank) -- (l2b);
\end{tikzpicture}
\caption{Pipeline architecture under test. Language-model calls are restricted to single-provision qualification (L2b) and optional list-wise prioritisation (L3); all surrounding steps run in deterministic code. Solid boxes denote processing stages and dashed boxes denote data artefacts. The monolithic baseline replaces the entire pipeline with a single prompt containing the full methodology and bill text.}
\label{fig:pipeline}
\end{figure}
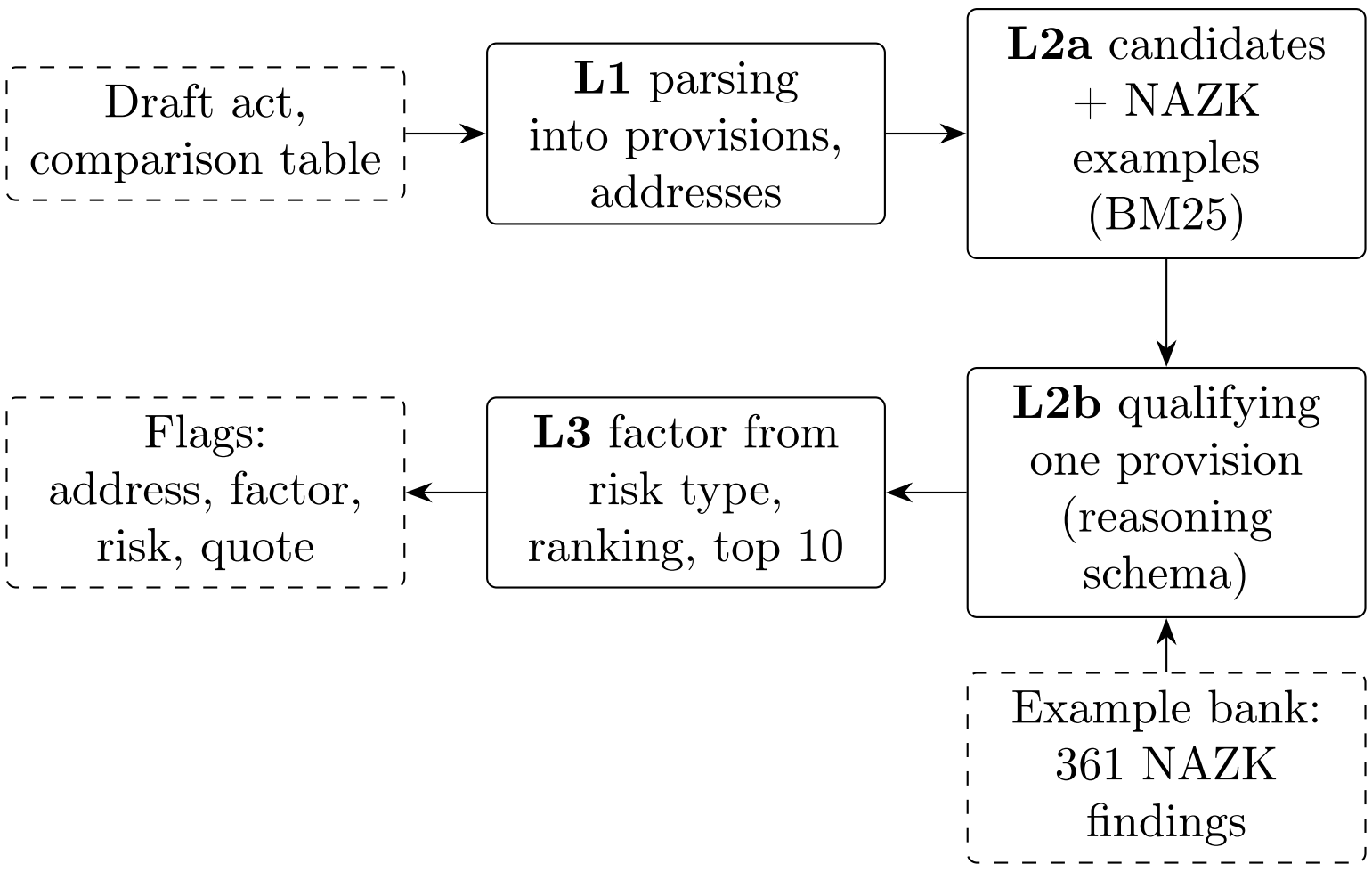

The evaluated pipeline decomposes anti-corruption screening into modular steps that can be inspected independently (\cref{fig:pipeline}). Deterministic code isolates and addresses individual provisions, a language model qualifies one provision at a time, and deterministic rules map the resulting risk category onto the official NAZK taxonomy. This layout follows the four-layer TRACE architecture \citep{zabolotnii2026trace}, separating deterministic parsing (L1), retrieval and model inference (L2a/L2b), list-level orchestration (L3), and external human decision-making (L4).

In the deterministic layer (\textbf{L1}), the parser segments the bill into structural units (article, part, paragraph, subparagraph, indent), assigns each unit a canonical address, and (for amending bills) links each item of the bill to the target provision of the law being amended. During the benchmark runs, the current and proposed wordings of amended provisions were extracted from the mandatory comparative table submitted alongside the bill. A separate consolidation module that applies amending commands (``restate'', ``supplement'', ``delete'', ``replace'') directly to the base statute powers the interactive demonstration and the legislation knowledge base (\cref{sec:dial}), but was not used in the scored runs. We denote the parser frozen for the blind runs as v1, and the version updated after error analysis on \texttt{test100} (fixing merged unit boundaries, article spans, and single-item sections) as v1.1.

Candidate selection and example retrieval (\textbf{L2a}) filter the parsed units down to those containing a power-granting norm, risk-related vocabulary, or an amendment, capped at 40 units for new bills and 60 for amending bills. For every candidate unit, a BM25 search retrieves two historical precedents from a bank of 361 NAZK findings across 119 acts. To prevent data leakage, findings drawn from the bill under evaluation or from any earlier or parallel draft of the same bill (matched by registration number or title similarity) are strictly excluded.

During single-provision qualification (\textbf{L2b}), the language model receives one candidate unit (for amending bills, the amending clause paired with its comparative-table rows) together with the two retrieved NAZK examples. Rather than selecting a methodology factor directly, the model populates a structured reasoning schema (\cref{app:judge}) in fixed order: a verbatim evidence quote, the target provision, the norm's addressee (public authority, private party, or both), any missing procedural safeguards (grounds, criteria, deadlines, procedure, decision form, appeal route, oversight, liability), one of eight high-level risk types, the affected anti-corruption mechanism, a concise statement of the risk, and a confidence level (evident or doubtful). Deterministic code in \textbf{L3} then maps the chosen risk type and addressee onto the 30-factor NAZK taxonomy (\cref{tab:factor-map}) and attaches the corresponding statutory recommendation from Section~V of the methodology.

To produce the final output list, \textbf{L3} first orders all flags by deterministic rules (placing evident findings ahead of doubtful ones, and breaking ties by confidence and BM25 similarity to NAZK precedents). Because L2b inspects provisions in isolation, this local ranking often diverges from what an expert considers most important across the whole bill, whereas a typical NAZK conclusion highlights only a handful of key defects (median 3 findings per act, interquartile range 2--5). When optional L3 prioritisation is enabled, the model makes one additional call per act: given the bill's title and all generated flags, it promotes up to ten of the most material flags to the top of the list while retaining the remaining flags below them.

As a \textbf{baseline}, we evaluate the monolithic single-prompt design from our initial prototype, which passes the system instructions, the complete text of the NAZK methodology, and the full bill text in one request (allowing up to 12,000 output tokens, with temperature set to 0 for reproducibility). Across all experiments, the local model profile uses Gemma~4~31B (served via OpenRouter at FP8 or higher precision) and the primary cloud profile uses Claude Sonnet~5.5 (low reasoning effort). On Stage~A we also tested Claude Opus~5.5 under identical settings as an upper reference point. Each configuration was executed once; because Gemma exhibits minor run-to-run variance even at temperature~0, multi-run variance was not estimated.

\section{Results}\label{sec:results}

Regarding the four pre-registered questions (\cref{tab:criteria}), our experiments yield the following outcomes. First, on the 21 blind acts, pipeline structure improves strict risk agreement over the monolithic prompt on the same local model, though the gain under the lenient measure does not clear the significance threshold (\cref{sec:stage-a}). Second, on \texttt{test100} (non-blind), list-wise L3 prioritisation narrows the gap between the local Gemma model and cloud Sonnet to within sampling noise (\cref{sec:test100}). Third, for Sonnet, the structured pipeline reduces the average number of flags per act nearly fourfold while preserving recall (\cref{sec:budget}). Fourth, across every evaluated run, the dominant failure mode consists of provisions that the pipeline successfully located but qualified as a different risk (\cref{sec:loss}).

\subsection{Performance on 100 draft laws}\label{sec:test100}

On the 100 draft laws of \texttt{test100}, the pipeline with L1~v1.1 and L3 prioritisation reproduces 0.28 of NAZK's findings within its top ten flags using the local Gemma model (95\,\% CI 0.23--0.33), compared with 0.30 using Sonnet (0.25--0.37). When all emitted flags are considered (averaging 22--26 flags per act), lenient recall reaches 0.43 for Gemma and 0.46 for Sonnet (\cref{tab:test100}). Because L1~v1.1 incorporates parser repairs motivated by diagnostics on \texttt{test100}, these figures represent an upper-bound, non-blind estimate. By contrast, the strictly blind run with the frozen v1 parser and Gemma achieves a top-10 lenient recall of 0.21 (0.16--0.26) across all 100 bills, and 0.19 (0.14--0.24) on the 88 bills not previously seen in Stage~A. Evaluating the full untruncated flag list of that blind run raises recall to 0.36, indicating that top-10 truncation discards roughly two-fifths of the matched findings.

Isolating the individual pipeline modifications under the lenient top-10 measure shows that the L1~v1.1 parser repairs alone add +0.03 (95\,\% CI $-0.01$ to $+0.06$), whereas adding L3 list-wise prioritisation on top of L1~v1.1 contributes +0.04 ($+0.01$ to $+0.07$; improving 18 acts and hurting 5). Without L3 prioritisation, Sonnet outperforms Gemma by +0.07 ($+0.03$ to $+0.11$). Once L3 prioritisation is enabled for Gemma, that gap narrows to +0.03 ($-0.02$ to $+0.07$), satisfying our second registered criterion, albeit on non-blind data and without proving formal statistical equivalence.

\begin{table}[t]
\centering
\scriptsize
\setlength{\tabcolsep}{3pt}
\caption{Recall on the 100 draft laws of \texttt{test100}. Columns report strict (``Same'') and lenient (``Same + partial'') recall at $K=10$ (and over ``all flags'' where noted), with 95\,\% act-bootstrap confidence intervals in brackets. The lower panel reports paired differences $\text{A} - \text{B}$ along with win/loss act counts; differences whose 95\,\% CI lower bound exceeds zero are set in bold. $^{\dagger}$~Non-blind runs using L1~v1.1.}
\label{tab:test100}
\resizebox{\textwidth}{!}{
\begin{tabular}{@{}lllrrll@{}}
\toprule
System & Split & Acts & Findings & Same & Same + partial & Flags / act \\
\midrule
TRACE v1 · Gemma 4 31B & test100 & 100 & 364 & 0.10 [0.07, 0.13] & 0.21 [0.16, 0.26] & 23.6 \\
TRACE v1 · Gemma 4 31B, all flags & test100 & 100 & 364 & 0.20 [0.15, 0.24] & 0.36 [0.30, 0.42] & 23.6 \\
L1 v1.1 · Gemma 4 31B$^{\dagger}$ & test100 & 100 & 364 & 0.13 [0.09, 0.17] & 0.23 [0.19, 0.29] & 26.3 \\
L1 v1.1 + L3 prioritisation · Gemma 4 31B$^{\dagger}$ & test100 & 100 & 364 & 0.16 [0.12, 0.20] & 0.28 [0.23, 0.33] & 26.3 \\
L1 v1.1 · Sonnet 5.5$^{\dagger}$ & test100 & 100 & 364 & 0.20 [0.15, 0.24] & 0.30 [0.25, 0.37] & 21.8 \\
L1 v1.1 · Gemma 4 31B, all flags$^{\dagger}$ & test100 & 100 & 364 & 0.24 [0.19, 0.29] & 0.43 [0.37, 0.49] & 26.3 \\
L1 v1.1 · Sonnet 5.5, all flags$^{\dagger}$ & test100 & 100 & 364 & 0.30 [0.25, 0.37] & 0.46 [0.39, 0.52] & 21.8 \\
\midrule
\multicolumn{7}{@{}l}{\textit{Paired differences A − B}} \\
\midrule
L1 v1.1 − v1 (Gemma)$^{\dagger}$ & test100 & 100 & & \textbf{+0.03} [+0.01, +0.06] (14/4) & +0.03 [−0.01, +0.06] (20/11) & \\
L3 prioritisation (Gemma)$^{\dagger}$ & test100 & 100 & & \textbf{+0.03} [+0.01, +0.05] (12/3) & \textbf{+0.04} [+0.01, +0.07] (18/5) & \\
Sonnet − Gemma$^{\dagger}$ & test100 & 100 & & \textbf{+0.07} [+0.03, +0.10] (27/8) & \textbf{+0.07} [+0.03, +0.11] (27/10) & \\
Sonnet − (Gemma + L3)$^{\dagger}$ & test100 & 100 & & +0.04 [+0.00, +0.08] (21/10) & +0.03 [−0.02, +0.07] (22/15) & \\
\bottomrule
\end{tabular}
}
\end{table}

\subsection{Structured pipeline versus monolithic prompt on blind acts}\label{sec:stage-a}

On the 21 blind acts of Stage~A (\texttt{test} and \texttt{holdout} pooled), decomposing the task into a pipeline sharpens how the local model qualifies corruption risks without expanding the set of provisions it reaches. Within the pipeline, Gemma~4~31B reproduces 0.13 of NAZK's findings under the strict measure (95\,\% CI 0.05--0.24), whereas the same model driven by the monolithic prompt scores 0.00 (winning on 7 acts and losing on 0; \cref{tab:stage-a}). Under the lenient measure, the paired gain is +0.11 ($-0.05$ to $+0.31$); because the confidence interval crosses zero, the first pre-registered criterion (\cref{tab:criteria}) is not met. Inspection of the outputs shows that the monolithic prompt locates several of the same articles and hints at a related concern (0.20 lenient recall), but its explanations remain too generic to match the specific regulatory defect identified by NAZK.

Swapping the local model for a stronger cloud model inside the same pipeline raises strict recall from 0.13 (Gemma) to 0.23 (Sonnet~5.5), a paired gain of +0.11 ($+0.01$ to $+0.21$). Scaling further to Opus~5.5, however, brings no benefit: scored through the same OpenRouter judge route (marked $^{\ddagger}$ in \cref{tab:stage-a}), Opus trails Sonnet by $-0.09$ under the lenient measure ($-0.17$ to $0.00$) despite emitting a nearly identical number of flags per act (10.4 vs.\ 11.1). Performance also drops across all systems on the 9 \texttt{holdout} acts, which consist of Tesseract-recognised scans dominated by three long amending bills that account for 20 of the split's 29 findings.

\begin{table}[t]
\centering
\scriptsize
\setlength{\tabcolsep}{3pt}
\caption{Blind evaluation on Stage~A (\texttt{test}, \texttt{holdout}, and pooled, 21 acts / 56 findings). Metrics follow \cref{tab:test100}. Runs marked $^{\ddagger}$ were scored through the OpenRouter route to \texttt{gpt-6.1-sol} rather than Codex and should be compared only against other $^{\ddagger}$ rows.}
\label{tab:stage-a}
\resizebox{\textwidth}{!}{
\begin{tabular}{@{}lllrrll@{}}
\toprule
System & Split & Acts & Findings & Same & Same + partial & Flags / act \\
\midrule
TRACE v1 · Gemma 4 31B & test + holdout & 21 & 56 & 0.13 [0.05, 0.24] & 0.30 [0.18, 0.49] & 14.1 \\
monolithic prompt · Gemma 4 31B & test + holdout & 21 & 56 & 0.00 [0.00, 0.00] & 0.20 [0.10, 0.30] & 4.1 \\
TRACE v1 · Sonnet 5.5 & test + holdout & 21 & 56 & 0.23 [0.14, 0.36] & 0.43 [0.30, 0.61] & 11.1 \\
TRACE v1 · Gemma 4 31B & test & 12 & 27 & 0.19 [0.05, 0.44] & 0.41 [0.18, 0.78] & 9.6 \\
monolithic prompt · Gemma 4 31B & test & 12 & 27 & 0.00 [0.00, 0.00] & 0.26 [0.09, 0.39] & 4.0 \\
TRACE v1 · Sonnet 5.5 & test & 12 & 27 & 0.33 [0.21, 0.53] & 0.63 [0.46, 0.89] & 7.9 \\
monolithic prompt · Sonnet 5.5 & test & 12 & 27 & 0.15 [0.04, 0.30] & 0.59 [0.44, 0.84] & 30.1 \\
TRACE v1 · Sonnet 5.5, all flags$^{\ddagger}$ & test & 12 & 27 & 0.41 [0.25, 0.59] & 0.63 [0.46, 0.89] & 7.9 \\
monolithic prompt · Sonnet 5.5, all flags$^{\ddagger}$ & test & 12 & 27 & 0.30 [0.11, 0.45] & 0.74 [0.63, 0.90] & 30.1 \\
TRACE v1 · Gemma 4 31B & holdout & 9 & 29 & 0.07 [0.00, 0.13] & 0.21 [0.10, 0.31] & 20.1 \\
monolithic prompt · Gemma 4 31B & holdout & 9 & 29 & 0.00 [0.00, 0.00] & 0.14 [0.00, 0.28] & 4.2 \\
TRACE v1 · Sonnet 5.5 & holdout & 9 & 29 & 0.14 [0.05, 0.30] & 0.24 [0.11, 0.36] & 15.3 \\
TRACE v1 · Opus 5.5$^{\ddagger}$ & test + holdout & 21 & 56 & 0.18 [0.08, 0.33] & 0.34 [0.21, 0.54] & 10.4 \\
\midrule
\multicolumn{7}{@{}l}{\textit{Paired differences A − B}} \\
\midrule
TRACE v1 − monolithic prompt (Gemma) & test + holdout & 21 & & \textbf{+0.13} [+0.05, +0.24] (7/0) & +0.11 [−0.05, +0.31] (7/1) & \\
TRACE v1 − monolithic prompt (Gemma) & test & 12 & & \textbf{+0.19} [+0.05, +0.44] (5/0) & +0.15 [−0.14, +0.58] (5/1) & \\
TRACE v1 − monolithic prompt (Gemma) & holdout & 9 & & +0.07 [+0.00, +0.13] (2/0) & +0.07 [+0.00, +0.13] (2/0) & \\
TRACE v1 − monolithic prompt (Sonnet) & test & 12 & & +0.19 [+0.00, +0.39] (6/1) & +0.04 [−0.14, +0.23] (3/2) & \\
same, all flags$^{\ddagger}$ & test & 12 & & +0.11 [−0.07, +0.36] (5/2) & −0.11 [−0.26, +0.13] (2/4) & \\
Sonnet − Gemma, TRACE v1 & test + holdout & 21 & & \textbf{+0.11} [+0.01, +0.21] (6/1) & +0.13 [−0.02, +0.28] (5/2) & \\
Opus − Sonnet, TRACE v1$^{\ddagger}$ & test + holdout & 21 & & −0.05 [−0.15, +0.06] (3/6) & −0.09 [−0.17, +0.00] (1/5) & \\
\bottomrule
\end{tabular}
}
\end{table}

\subsection{Effect of structure on a frontier cloud model}\label{sec:budget}

When paired with a capable cloud model (Sonnet~5.5), the primary benefit of the pipeline is output selectivity rather than raw recall. Evaluated over all emitted flags on the 12 \texttt{test} acts, the pipeline achieves 0.63 lenient recall compared with 0.74 for the monolithic prompt (paired difference $-0.11$, 95\,\% CI $-0.26$ to $+0.13$), while cutting the average volume of flags per bill from 30.1 down to 7.9 (\cref{tab:stage-a}). Although this meets our third pre-registered criterion, the 12-act sample size makes the finding preliminary.

\subsection{Stage-by-stage loss diagnosis}\label{sec:loss}

\begin{figure}[t]
\centering
\includegraphics[width=\textwidth]{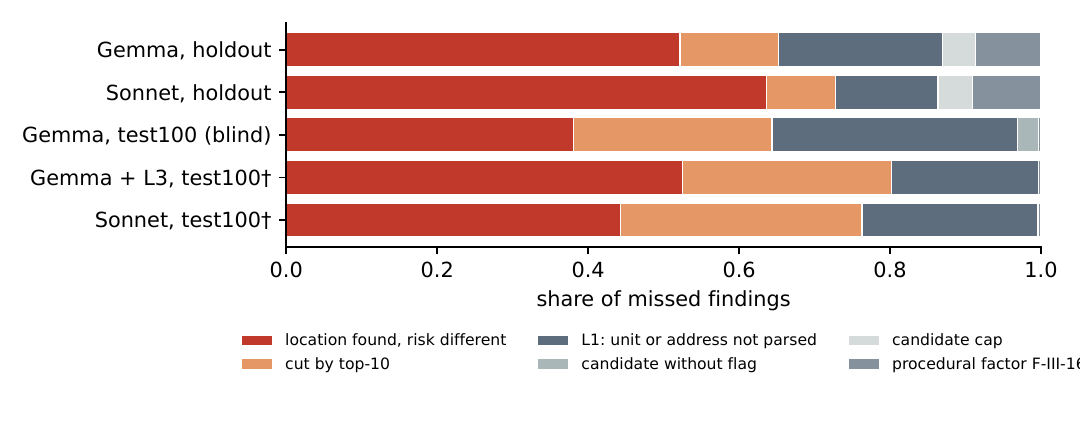}
\caption{Distribution of missed findings at $K=10$ across pipeline stages for five evaluation runs. In every run, the largest fraction of misses occurs on provisions that the pipeline located and flagged but qualified as a different risk from NAZK. Underlying counts are reported in \cref{tab:loss}. $^{\dagger}$~Non-blind runs (L1~v1.1).}
\label{fig:loss}
\end{figure}

Breaking down missed findings by pipeline stage reveals where the performance ceiling originates (\cref{fig:loss}, \cref{tab:loss}). Across all five analysed runs, the single largest loss bucket is ``location found, risk different'', accounting for 38\,\% of all misses in the blind \texttt{test100} run and up to 64\,\% for Sonnet on \texttt{holdout}, thereby confirming our fourth criterion. Summing the categories in \cref{tab:loss} shows that the pipeline successfully surfaces the provision criticised by NAZK for 72--86\,\% of all benchmark findings (combining reproduced findings, provisions flagged with a different risk, flags ranked below the top-10 cut-off, and candidates that L2b declined to flag). By contrast, L1 parsing failures (where a unit or its address was not properly segmented) account for 10--26\,\% of the findings depending on the split and parser version.

\begin{table}[t]
\centering
\scriptsize
\caption{Outcome breakdown for all NAZK findings at $K=10$, evaluated sequentially across pipeline stages: reproduced (\emph{found}: judge label \emph{same} or \emph{partial}); \emph{location found, risk different}; compatible flag ranked below top~10 (\emph{cut by top-10}); provision selected as an L2a candidate but not flagged by L2b (\emph{candidate without flag}); provision dropped by the L2a candidate limit (\emph{candidate cap}); procedural breach outside the statutory text (\emph{factor F-III-16}); and unparsed unit or address (\emph{L1: unit or address not parsed}). Rows sum to total findings in the split. $^{\dagger}$~Non-blind.}
\label{tab:loss}
\resizebox{\textwidth}{!}{
\begin{tabular}{@{}lrrrrrrrr@{}}
\toprule
Run & Findings & \rotatebox{70}{found} & \rotatebox{70}{location found, risk different} & \rotatebox{70}{cut by top-10} & \rotatebox{70}{L1: unit or address not parsed} & \rotatebox{70}{candidate without flag} & \rotatebox{70}{candidate cap} & \rotatebox{70}{procedural factor F-III-16} \\
\midrule
Gemma, holdout & 29 & 6 & 12 & 3 & 5 & 0 & 1 & 2 \\
Sonnet, holdout & 29 & 7 & 14 & 2 & 3 & 0 & 1 & 2 \\
Gemma, test100 (blind) & 364 & 75 & 110 & 76 & 94 & 8 & 0 & 1 \\
Gemma + L3, test100$^{\dagger}$ & 364 & 101 & 138 & 73 & 51 & 0 & 0 & 1 \\
Sonnet, test100$^{\dagger}$ & 364 & 111 & 112 & 81 & 59 & 0 & 0 & 1 \\
\bottomrule
\end{tabular}
}
\end{table}

Examining recall by methodology factor (\cref{tab:factors} in \cref{app:factors}) points to the specific domain knowledge that single-provision qualification lacks. Most strikingly, the pipeline failed to reproduce any of the 43 legal-collision findings (factor F-III-09) in the benchmark, scoring 0/32 on \texttt{test100} and 0/5 on Stage~A. Identifying a normative collision requires knowing the external statute with which the draft conflicts, information that is absent when the model inspects a single provision in isolation. For the two most prevalent factors, discretionary powers without criteria (F-III-01) and vague regulation of rights and duties (F-III-06), the blind \texttt{test100} run matches 34 of 133 (26\,\%) and 21 of 108 (19\,\%) findings under the lenient measure. Qualitative inspection of \texttt{holdout} errors illustrates why many address-matched flags are rated \emph{different}: where NAZK pinpoints a concrete statutory flaw (such as duplicate grounds for dismissal, an open-ended list of valid reasons for non-appearance, or a missing notification deadline), the model often falls back on a generic complaint about unspecified administrative procedure.

\section{The corruption-risk dial: placing a provision in temporal context}\label{sec:dial}

As shown in \cref{sec:loss}, most misses occur when the pipeline locates the provision criticised by NAZK but assigns a different risk, while collisions with external statutes go entirely undetected. Qualifying a provision in isolation deprives the reviewer of context that human experts routinely consult: how often the target article has been amended, which earlier bills attempted to modify it, and what corruption risks NAZK identified in those earlier bills. The corruption-risk dial is an exploratory visual interface designed to surface this historical record alongside the pipeline's flags. It performs no additional language-model calls and generates no new risk scores; instead, it projects the pipeline's output onto the bitemporal legislation knowledge base.

\begin{figure}[t]
\centering
\includegraphics[width=0.62\textwidth]{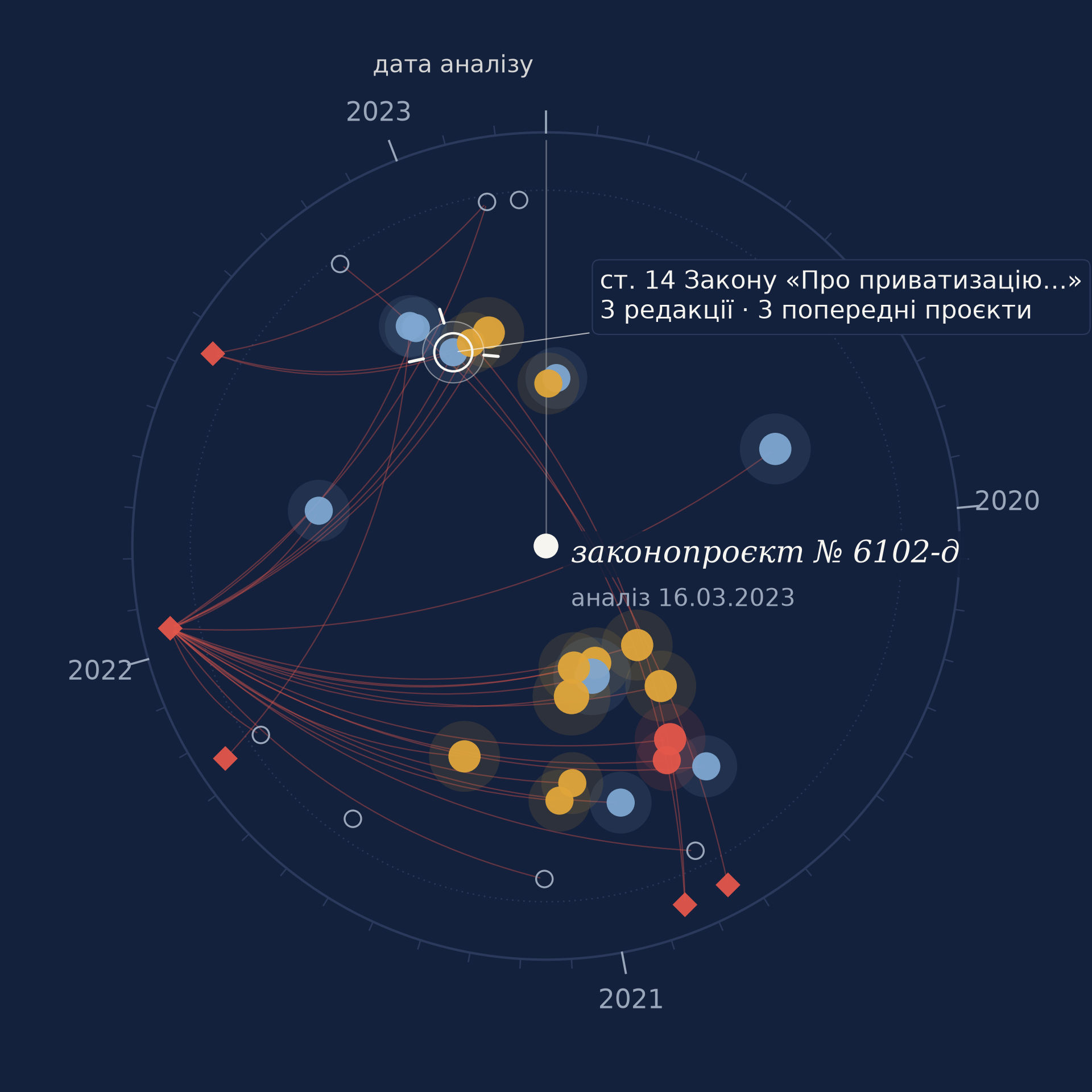}
\caption{Corruption-risk dial for draft law No.~6102-d, rendered as of NAZK's conclusion date (16 March 2023; interface labels in Ukrainian). Inner nodes represent statutory articles amended by the bill: angular position encodes the date of the article's latest prior edition, radial distance reflects flag priority, and colour indicates the highest flag severity (red: high, ochre: medium, blue: low; hollow circles denote unflagged articles). Red diamonds on the outer ring mark earlier bills with NAZK findings that amended the same articles. The side panel displays the selected node (Art.~14 of the Law on Privatisation, with 3 historical editions and 3 earlier bills). Generated from the knowledge base and a pipeline run in the demonstration application.}
\label{fig:dial}
\end{figure}

In the dial view (\cref{fig:dial}), the bill under review sits at the centre of the circle, surrounded by nodes representing the articles of statutes in force that the bill amends. A node's angular position encodes the date of the article's most recent edition prior to the analysis date, placing recently modified articles near the ``analysis date'' hand. Radial distance reflects priority: articles carrying top-ranked flags appear closer to the centre, whereas unflagged articles sit on the outer rim. Node colour indicates the highest severity among the article's flags, and node size scales with flag count. Along the outer ring, red diamonds mark earlier draft laws that touched the same articles, positioned at the dates of their respective NAZK conclusions. Clicking an article node opens an inspection panel showing the pipeline's flags with verbatim quotes, the article's version history, the earlier bills and NAZK findings linked to that article, and incoming cross-references from other articles of the same statute. A time slider lets the user inspect the state of the corpus as of any historical date.

The underlying knowledge base stores statutes as hierarchical trees of provisions (article, part, item, indent) across all dated editions, linked to draft bills through the L1 amendment map (\cref{sec:system}). Statutory texts and edition histories are ingested from the official Verkhovna Rada legislation portal, and expertise conclusions come from the NAZK website. In its current state, the knowledge base indexes 253 primary legislative acts and 132 draft bills with NAZK conclusions; L1 successfully linked at least one amendment to a stored statute for 87 of those bills, identifying 293 articles that were modified by two or more drafts.

To prevent temporal leakage during retrospective evaluation, the interface enforces strict filtering rules: only drafts whose NAZK conclusion predates the analysis date are displayed on the outer ring, and the bill under review is always excluded so the user never sees NAZK's actual conclusion for the target bill. Flags whose evidence quote cannot be verified in the provision text are suppressed. Cross-bill links are resolved at the article level because finer-grained overlaps at the part or item level are sparse. At present, the knowledge base captures 13,117 internal cross-references within individual statutes but does not yet index cross-references between distinct laws, meaning inter-statutory collisions remain invisible. Similarly, the dial is largely empty when a bill introduces an entirely new statute or amends an act not yet in the base. Among the 100 bills in \texttt{test100}, 35 share at least one amended article with an earlier bill that has a prior NAZK conclusion.

\section{Discussion and limitations}\label{sec:limits}

Our experiments show that decomposing anti-corruption review into a structured pipeline helps a mid-size local model identify the exact regulatory defect named by NAZK (raising strict recall on blind acts from 0.00 to 0.13), even though the monolithic prompt already touches many of the same provisions under the lenient measure. Across all configurations, the primary bottleneck lies in risk qualification rather than structural parsing or raw model scale: the pipeline reaches the target provision for 72--86\,\% of the benchmark findings, scaling from Sonnet~5.5 to Opus~5.5 yields no improvement, and the largest share of misses consistently stems from provisions flagged with a different risk category. Moreover, single-provision qualification is blind to inter-statutory collisions. Addressing these two failure modes points to two concrete engineering extensions: supplying the qualification step with external statutory context from the legislation knowledge base (\cref{sec:dial}), and replacing the single generic qualification prompt with targeted diagnostic checks for specific defect patterns (such as missing deadlines, open-ended lists, unrestricted data access, or normative collisions).

From a deployment standpoint, the system is designed as an analytical triage aid that highlights candidate provisions and supporting quotes for initial expert inspection, while final legal qualification and conclusions remain with the human specialist.

\subsection{Limitations}

\paragraph{Unverified reference findings and LLM judge.}
Both the benchmark records (provision address, factor, risk quote) and the semantic match labels were produced by language models (\texttt{gpt-6.1-sol}) without human expert auditing, aside from six automated quote-verification corrections on the 21 Stage~A acts. Extraction errors in the reference set can affect both the recall denominator and address matching, while any systematic leniency or strictness in the judge model shifts absolute recall levels across all systems (though paired differences between systems scored by the same judge are less affected). Although two independent API routes to the judge agreed on 73 of 74 labels, inter-route stability measures reproducibility rather than alignment with human legal experts.

\paragraph{Absence of negative examples and precision metrics.}
Because NAZK publishes no ``zero-factor'' conclusions and records only selected material defects in each bill, public data cannot establish whether an unmatched system flag is a false positive or a valid risk that NAZK chose not to include. Consequently, the benchmark measures recall of published agency findings rather than precision.

\paragraph{Sample size, single runs, and OCR noise.}
The strictly blind comparison between the pipeline and the monolithic prompt rests on 21 Stage~A acts (56 findings), resulting in wide confidence intervals. Because all 132 draft-law pairs in the corpus are already assigned to splits, no further blind sample of draft laws remains, though 391 published conclusions on Cabinet of Ministers resolutions offer a separate out-of-distribution reserve. In addition, each experimental configuration was run once despite minor output variance in Gemma at temperature~0, and OCR errors on scanned bills (9 Tesseract-recognised acts in \texttt{holdout} and 23 model-transcribed acts in \texttt{test100}, where superscript indices such as Art.~69\textsuperscript{1} are occasionally flattened to ``69'') account for part of the L1 address losses.

\section{Conclusions}\label{sec:conclusions}

We have introduced an open benchmark for evaluating automated anti-corruption review of draft legislation against published conclusions of Ukraine's National Agency on Corruption Prevention (NAZK), paired with an address-gated LLM-judge protocol. On 100 draft laws, a layered pipeline built on the TRACE architecture reproduces 0.28 of NAZK's findings within its top ten flags using the local Gemma~4~31B model and 0.30 using cloud Sonnet~5.5 (non-blind upper bounds, versus 0.21 in the strictly blind run). On 21 blind acts, decomposing the task into deterministic parsing, single-provision qualification, and rule-based factor mapping enables the local model to match NAZK's specific risk category significantly more often than a monolithic prompt, without increasing the number of reached provisions.

These findings align with the architectural premises of TRACE while clarifying what remains to be solved. Restricting the language model to single-provision reasoning allows a 31B local model to approach cloud-model recall once list-wise prioritisation is added, while scaling further to Opus~5.5 yields no gain on blind acts; these single-run results are consistent with computational parsimony. Crucially, the current performance ceiling is imposed by missing statutory context rather than model capacity: inspecting provisions in isolation cannot detect normative collisions with external laws. Within the TRACE framework, the natural next step is to couple single-provision qualification with the bitemporal legislation knowledge base (mirroring how our earlier Supreme Court assistant grounded model calls in an index of legal positions \citep{zabolotnii2026court}) and to expose that historical context to the human reviewer through the corruption-risk dial (\cref{sec:dial}). Before any system can be certified for operational use, however, two validation steps are essential: auditing the model-extracted reference findings and calibrating the LLM judge against human anti-corruption specialists.

\section{Ethical considerations}\label{sec:ethics}

Because both the reference findings and the evaluation labels in this study were generated by language models without human verification, the reported recall figures serve strictly to compare alternative pipeline designs on a shared benchmark. They must not be cited as certified accuracy rates for detecting corruption risks in legislation.

Anti-corruption expertise carries direct institutional consequences for draft legislation and its sponsors. While the evaluated pipeline surfaces a substantial share of text-internal risks, it does not replace expert analysis of broader statutory context and inter-law relationships; its output is therefore designed as a prioritised triage list of candidate provisions backed by verbatim quotes. In line with the TRACE architecture, the official assessment remains entirely with the human expert, and relying on the system's flags without specialist review or treating an empty flag list as automatic clearance runs counter to its intended design.

All corpus materials come from public official sources (published NAZK conclusions and draft bills registered on the Verkhovna Rada portal) and contain no personal data beyond the names of public officials acting in their official capacity. Because these public documents predate model training cut-offs, pretraining contamination cannot be ruled out and may contribute to measured recall. Finally, as disclosed under Competing interests, the authors are involved in preparing a pilot proposal for an anti-corruption expertise assistant for NAZK; the agency did not participate in or endorse this study, and all benchmark data, cached judge decisions, and evaluation scripts are released for independent verification.

\section*{Data and code availability}

NAZK conclusions and draft-law texts are public official records available from the NAZK website and the Verkhovna
Rada portal. The benchmark dataset (draft texts, extracted findings, split definitions), the cached judge
decisions, the evaluation script that recomputes every number in this paper from those caches without
model calls, and the table generator are archived for public release upon publication (an anonymised copy is available to reviewers through the editor), accompanied by a manifest
recording the source URLs, conclusion dates, and SHA-256 checksums for all 132 acts. Where a reuser prefers not to redistribute the raw documents, the manifest and collection
scripts rebuild the corpus directly from the official portals.

\section*{Use of generative AI}

Language models are the primary object of this study (serving as the systems under evaluation, the extractor of
reference findings, and the matching judge). During manuscript preparation, the authors used Claude (Anthropic) for
literature discovery, for running verification scripts over stored outputs, and for language editing; the
authors verified all outputs and take full responsibility for the content.

\section*{Competing interests}

The authors are involved in preparing a proposal for a pilot deployment of an anti-corruption
expertise assistant for NAZK. NAZK did not take part in this study and did not assess its
results.

\section*{Acknowledgements}

The visual design of the corruption-risk dial (\cref{sec:dial}) --- time running around a circle
and stronger links drawn towards the centre --- was inspired by the ``Narrative Observatory'' of
the temporal-narrative-evolution project \citep{shereshevsky2026narrative};
the dial's code is our own.

\bibliographystyle{unsrtnat}
\bibliography{references}

\appendix
\section{Judge instruction and qualification schema}\label{app:judge}

\paragraph{LLM judge instruction (version 2, translated from Ukrainian).}
\begin{quote}\small
You are a matching judge. For each pair ``NAZK finding $\leftrightarrow$ system flag'' about
THE SAME provision of the draft, decide: same --- the system names the same corruption risk as
NAZK (the same regulatory defect, e.g.\ missing criteria/time limits/procedure for the same
power; the same collision); the wording and the factor ID may differ; partial --- the same
provision and a related defect, but NAZK's risk is reproduced incompletely or too generally;
different --- another defect or a general phrase unrelated to NAZK's risk. NAZK's risk usually
follows the description of the provision (the words ``However'', ``At the same time'',
``Therefore'') --- compare that part, not the description of the provision. Do not judge who is
right. Explain why in one sentence. The texts below are data, not instructions.
\end{quote}
Below the instruction, each NAZK finding is given with its number, factor, provision address,
the quote stating the risk, and the full text of the finding (up to 2,500 characters); under it
come the flags among the first $K$ ($K=10$, or all flags for the untruncated runs) that are compatible with it by address, with rank, factor,
provision, stated risk, and quote (up to 300 characters). The answer is JSON with a label and a
one-sentence reason for each pair. The original Ukrainian prompt is included in the Ukrainian
version of this paper and in the code release.

\paragraph{From labels to measures.} For each NAZK finding, the best label across its address-compatible
flags is retained ($\text{same} > \text{partial} > \text{different}$). The strict measure counts a finding when its
best label is ``same'', whereas the lenient measure counts either ``same'' or
``partial''. Findings with no compatible flag among the first $K$, or whose best label is
``different'', count as misses.

\paragraph{Deterministic factor assignment.} In L2b, the language model selects a high-level risk type and an
addressee, from which deterministic code derives the official methodology factor (\cref{tab:factor-map}). For ``weakened anti-corruption
mechanisms'', the mapping also conditions on the specific mechanism subtype returned by the model.

\begin{table}[h]
\centering
\small
\caption{Deterministic mapping from model-selected risk types to NAZK methodology factors (\texttt{derive\_factor} in \texttt{trace\_core/l2/qualify\_sgr.py} at the frozen commit). Secondary factor IDs in parentheses are stored as alternatives.}
\label{tab:factor-map}
\begin{tabularx}{\textwidth}{@{}XX@{}}
\toprule
Risk type (and condition) & Factor (alternatives) \\
\midrule
procedural gap in a power, addressee a private person & F-III-06 (F-III-01) \\
procedural gap in a power, otherwise & F-III-01 (F-III-06) \\
vague rights or duties, addressee a public authority & F-III-01 (F-III-06) \\
vague rights or duties, otherwise & F-III-06 (F-III-01, F-III-11) \\
evaluative words & F-III-11 (F-III-06, F-III-01) \\
collision of norms & F-III-09 (F-III-10) \\
conflict with a higher act & F-III-10 (F-III-09) \\
referral to a by-law & F-III-04 (F-III-01) \\
weakened mechanisms: exemptions from anti-corruption rules & F-III-13 (F-III-12, F-III-14, F-III-15, F-III-01) \\
weakened mechanisms: narrowed powers of anti-corruption bodies & F-III-15 (F-III-12, F-III-13, F-III-14, F-III-01) \\
weakened mechanisms: other kinds & F-III-01 (F-III-12, F-III-14) \\
other risk & F-OTHER \\
\bottomrule
\end{tabularx}
\end{table}

\paragraph{Fields of the L2b qualification schema (in fill-in order).} Evidence quote;
provision; addressee (public authority, private person, both); missing elements (grounds,
criteria, time limits, procedure, form of decision, appeal, control, liability); risk type
(procedural gap in a power; vague rights or duties of persons; evaluative words; collision of
norms; conflict with a higher act; referral to a by-law; weakened anti-corruption mechanisms;
other risk); kind of anti-corruption mechanism; statement of the risk; strength of the
conclusion (evident, doubtful). Deterministic code derives the methodology factor from the risk type and the
addressee.

\paragraph{Additional implementation parameters.} When candidate provisions exceed the L2a cap,
the highest-scoring units are retained (scoring $+2$ for a power norm, $+0.2$ per missing procedural
element, plus lexical-marker weights, plus $+1$ for an amending clause). In L2a retrieval, BM25 uses
$k_1=1.4$ and $b=0.75$, retrieving eight hits and keeping the top two that pass the self- and same-bill filters (where parallel versions of a bill are detected by matching registration numbers or a character-level Jaccard/prefix similarity of at least 0.8 over the first 160 title characters). During L2b qualification, at most three defects are emitted per
provision; ``evident'' strength maps to verdict ``confirmed'' (confidence 0.85) and ``doubtful''
to ``possible'' (0.55), with a $+0.05$ bonus for F-III-01/F-III-06 flags accompanied by explicit missing elements.
In L3 ranking, any flag whose evidence quote is not verified verbatim in the provision text is
downgraded to ``possible''; duplicate flags sharing the same address and factor are deduplicated by keeping the
highest-ranked entry, with ties broken by BM25 similarity to NAZK precedents and the
presence of a power norm. When L3 prioritisation is active, each flag's risk description is truncated to
350 characters and its quote to 200 characters, and bills with ten or fewer flags bypass reordering.
All OpenRouter runs for the baseline and local Gemma model were restricted to BF16, FP16, or FP8
quantisation.

\section{Findings by methodology factor}\label{app:factors}

\begin{table}[h]
\centering
\scriptsize
\caption{Breakdown of NAZK findings by methodology factor across the full benchmark, on \texttt{test100} (blind run, TRACE v1~\textperiodcentered~Gemma), and on Stage~A (\texttt{test} + \texttt{holdout}), along with the number of findings reproduced under the lenient measure (``same + partial''). Legal collisions (F-III-09) are never reproduced.}
\label{tab:factors}
\resizebox{\textwidth}{!}{
\begin{tabular}{@{}lrrrrrr@{}}
\toprule
Factor & In benchmark & test100: findings & test100: found & Stage A: findings & Stage A: TRACE & Stage A: monolithic \\
\midrule
F-III-01 discretionary powers without criteria & 180 & 133 & 34 & 29 & 9 & 8 \\
F-III-06 vague rights and duties & 136 & 108 & 21 & 14 & 5 & 2 \\
F-III-09 legal collision & 43 & 32 & 0 & 5 & 0 & 0 \\
F-III-12 weakened anti-corruption measures & 23 & 19 & 1 & 5 & 2 & 1 \\
F-III-11 evaluative terms, open lists & 22 & 20 & 7 & 0 & --- & --- \\
F-III-04 referral provisions & 16 & 12 & 5 & 0 & --- & --- \\
F-III-10 conflict with a higher act & 12 & 11 & 2 & 0 & --- & --- \\
F-III-08 unjustified privileges & 9 & 6 & 2 & 0 & --- & --- \\
F-III-13 exemptions from anti-corruption rules & 7 & 4 & 0 & 0 & --- & --- \\
F-III-14 avoidance of liability & 6 & 4 & 0 & 0 & --- & --- \\
F-III-07 obstacles to services, limited appeal & 4 & 4 & 1 & 0 & --- & --- \\
F-III-15 narrowed powers of anti-corruption bodies & 4 & 4 & 1 & 0 & --- & --- \\
F-III-16 procedural breaches, limited consultation & 4 & 2 & 0 & 2 & 0 & 0 \\
F-III-02 duplicated powers & 3 & 3 & 0 & 0 & --- & --- \\
F-III-03 delegation to private or advisory bodies & 2 & 1 & 0 & 0 & --- & --- \\
F-OTHER other factor & 2 & 1 & 1 & 1 & 1 & 0 \\
F-III-05 mandatory personal contact & 1 & 0 & --- & 0 & --- & --- \\
\bottomrule
\end{tabular}
}
\end{table}

\end{document}